\documentclass[11pt]{article}

\usepackage[preprint]{acl}

\usepackage{times}
\usepackage{latexsym}
\usepackage{amsmath}
\usepackage{amsfonts}
\usepackage{dsfont}
\usepackage{booktabs}
\usepackage{multirow}
\usepackage{url}

\usepackage[T1]{fontenc}

\usepackage[utf8]{inputenc}

\usepackage{microtype}

\usepackage{inconsolata}

\usepackage{graphicx}

\title{Exploring Fine-Tuning for In-Context Retrieval and Efficient KV-Caching in Long-Context Language Models}

\author{Francesco Maria Molfese\thanks{Work done at Amazon AGI.}$^\diamondsuit$ \quad
  Momchil Hardalov$^\spadesuit$ \quad
  Rexhina Blloshmi$^\spadesuit$ \\
  \textbf{Bill Byrne}$^\spadesuit$ \quad
  \textbf{Adri\`a de Gispert}$^\spadesuit$ \\[0.5em]
  $^\diamondsuit$Sapienza University of Rome \quad $^\spadesuit$Amazon AGI  \\[0.3em]
  \texttt{molfese@diag.uniroma1.it} \\
  \texttt{\{momchilh, blloshmi, willbyrn, agispert\}@amazon.com}
}

\begin{document}
\maketitle

\begin{abstract}
With context windows of millions of tokens, Long-Context Language Models (LCLMs) can encode entire document collections, offering a strong alternative to conventional retrieval-augmented generation (RAG).
However, it remains unclear whether fine-tuning strategies can improve long-context performance and translate to greater robustness under KV-cache compression techniques.
In this work, we investigate which training strategies most effectively enhance LCLMs' ability to identify and use relevant information, as well as enhancing their robustness under KV-cache compression.
Our experiments show substantial in-domain improvements, achieving gains of up to +20 points over the base model.
However, out-of-domain generalization remains task dependent with large variance -- LCLMs excels on finance questions (+9 points), while RAG shows stronger performance on multiple-choice questions (+6 points) over the baseline models.
Finally, we show that our fine-tuning approaches bring moderate improvements in robustness under KV-cache compression, with gains varying across tasks.
\end{abstract}

\section{Introduction}\label{sec:introduction}
Long-Context Language Models~(LCLMs) have demonstrated remarkable performance across diverse benchmarks through continual pre-training and context extension techniques~\cite{peng2023yarnefficientcontextwindow, su2023roformerenhancedtransformerrotary, xiong-etal-2024-effective, gao-etal-2025-train}.
The main goal is to enable the encoding of increasingly longer documents, potentially encompassing entire knowledge bases and eliminating the need for retrieval pipelines~\cite{lee2024longcontextlanguagemodelssubsume}. 
This motivation stems from the fact that retrieval-augmented generation (RAG), while effective as the de-facto standard for knowledge-intensive tasks, requires storing and managing comprehensive knowledge indexes while maintaining robust retrieval capabilities. 
However, despite the significant progress in long-context modeling, current literature demonstrates that standard RAG systems continue to outperform LCLMs across various tasks~\cite{ li-etal-2024-retrieval, lee2024longcontextlanguagemodelssubsume, bai-etal-2024-longbench, xu2024retrievalmeetslongcontext, jin2024longcontextllmsmeetrag,li2025longcontextvsrag}.
Recent works have explored the effects of supervised fine-tuning~(SFT) on LCLMs through question answering over long-context inputs~\cite{zhang2024raftadaptinglanguagemodel, jin2024longcontextllmsmeetrag, qiu-etal-2025-eliciting}, yet these studies typically train and evaluate on in-domain data, falling short in assessing their out-of-domain generalization and the factors contributing to the model performance.
Moreover, little attention has been devoted to understanding which training techniques yield the best performance under KV-cache compression~\cite{NEURIPS2023_6ceefa7b,pmlr-v202-sheng23a,NEURIPS2024_fd070571}, a critical consideration given the cost of storing complete key-value caches.

We argue that to effectively replace RAG systems, models must learn to behave like in-context learning RAG systems: attending primarily to relevant information within the input context while disregarding irrelevant content.
This motivates our research question: \textit{Does fine-tuning LCLMs to selectively attend to relevant information improve in-context retrieval performance and enable more efficient inference?}

Given that current SFT techniques are not able to achieve consistent performance gains against standard RAG pipelines~\cite{qiu-etal-2025-eliciting}, we turn to Group Relative Policy Optimization~\cite[GRPO]{shao2024deepseekmathpushinglimitsmathematical}. 
We argue that while SFT constrains the models to specific training paths, GRPO enables them to explore diverse strategies to discover which information is truly relevant to answering input questions.
We therefore explore GRPO-based approaches with verifiable reward functions, from simple answer-only objectives to complex reasoning with LLM-as-a-Judge evaluation, comparing their performance against RAG pipelines in both in- and out-of-domain scenarios.

Our findings show that certain strategies surpass RAG on in-domain benchmarks, though out-of-domain generalization remains task dependent with large variance.
Finally, while we show that fine-tuning brings moderate improvements in robustness under KV-cache compression, our analysis reveals minimal gains in attention-based document ranking, exhibiting negligible correlation with task performance.\footnote{To ensure reproducibility and encourage further research in LCLMs, we release our code at \url{https://github.com/amazon-science/icr-kv-caching-long-context-llms}.}

\section{Related Work}\label{app:related-work}
\paragraph{Long Context vs. RAG.}
Recent work has explored the trade-offs between LCLMs and RAG, revealing inconsistent findings across different settings~\cite{li2025longcontextvsrag}.
While some studies suggest that retrieval helps models with smaller context windows (4k) but not longer ones (16k-32k)~\cite{bai-etal-2024-longbench}, others demonstrate that RAG consistently outperforms LCLMs at context lengths up to 32k~\cite{xu2024retrievalmeetslongcontext, li-etal-2024-retrieval, yu2024defenserageralongcontext, xu2025chatqa2bridginggap}.
Conversely, \citet{bai-etal-2025-longbench} show that LCLMs can perform better without RAG at context lengths up to 128k, while \citet{jiang2024longragenhancingretrievalaugmentedgeneration} demonstrate that RAG benefits from longer retrieval units.
These mixed findings indicate ongoing debate about whether LCLMs can effectively replace RAG or whether retrieval remains essential for long-context understanding.

\paragraph{Fine-tuning LCLMs.}
Several approaches have attempted to improve long-context capabilities through fine-tuning on question answering tasks with distractor documents~\cite{zhang2024raftadaptinglanguagemodel, jin2024longcontextllmsmeetrag, qiu-etal-2025-eliciting}. 
RAFT~\cite{zhang2024raftadaptinglanguagemodel} trains models to ignore irrelevant documents when answering questions, while \citet{qiu-etal-2025-eliciting} use challenging distractors from strong retrievers and require models to generate both document IDs and content.
However, both focus exclusively on supervised fine-tuning (SFT) and evaluate primarily on in-domain RAG benchmarks, limiting insights into out-of-domain generalization and whether models genuinely improve attention to relevant content.
\citet{jin2024longcontextllmsmeetrag} explore answer-only training and reasoning distillation from stronger models, demonstrating that SFT with reasoning improves performance~\citep{openai2025,gemini2025,amazon2025nova2,olmo2025olmo}, though their analysis remains limited to in-domain settings.

\paragraph{} Our work extends this line of research in three key directions: (1) we explore reinforcement learning (GRPO) beyond standard SFT, with and without reasoning objectives; (2) we evaluate systematically on both in-domain and out-of-domain benchmarks; and (3) we analyze the mechanisms underlying performance improvements through attention-based document ranking and KV-cache compression robustness, rather than focusing solely on downstream task metrics.

\section{Methodology}\label{sec:methodology}

\paragraph{Training Data.}\label{sec:methodology-data}
To improve the in-context retrieval capabilities of the LCLMs,
we generate training data containing a sparse set of relevant documents within the input context, with the vast majority of information consisting of hard negatives that encourage the model to attend selectively to relevant content.
We construct our training data from two open-domain question answering benchmarks: HotpotQA~\cite{yang-etal-2018-hotpotqa} and 2WikiMultihopQA~\cite{ho-etal-2020-constructing}.
Each instance in these benchmarks includes an annotated list of (up to 2) relevant passages from Wikipedia required to answer the question.

We employ a strong retriever to fetch additional Wikipedia passages related to each question and concatenate them with the gold relevant passages.
This creates a challenging setting where the majority of retrieved information may be topically related to the question but not necessary for answering it~\cite{qiu-etal-2025-eliciting}.
Following \citet{karpukhin-etal-2020-dense}, we parse Wikipedia by dividing each article into 100-word passages, ensuring no length bias during training (the gold passages provided in these datasets also average approximately 100 words).
To prevent positional bias during training, we randomly shuffle both the relevant passages and hard negatives within each context.
Then, for each document $d_i$ in the context, we associate it with a unique ID using a special tag (i.e.,~``[DOC i]''), in order to distinguish clear document boundaries.

Additional details and examples about training data can be found in Appendix \ref{app:training-data}.

\paragraph{Training Strategies.}\label{sec:methodology-training}
Given that current SFT techniques for eliciting in-context learning capabilities of LCLMs have shown limited performance improvements against standard RAG pipelines~\cite{jin2024longcontextllmsmeetrag, qiu-etal-2025-eliciting}, we hypothesize that LCLMs require exploration during training to discover which information is truly relevant to answering input questions.
To enable this exploration, we employ reinforcement learning with 
GRPO, which allows the model to sample multiple outputs and learn from group-relative advantages.

Given a question $q$ and context $c$ containing passages $\{d_1, \ldots, d_p\}$, we train a policy model $\pi_\theta$ to generate a response $y$ that identifies relevant documents and produces the correct answer $a^*$.
Let $\mathcal{D}^* \subseteq \{1, \ldots, p\}$ denote the set of gold-standard relevant document indices.
We design five verifiable reward functions $R(y, q, c)$ to guide model training, each encouraging different aspects of relevant information extraction and reasoning.

\paragraph{Reward Functions.}
To encourage models to develop selective attention capabilities towards relevant information, we design five reward functions that progressively increase in complexity and constraint. 
We hypothesize that simpler objectives (i.e., answer correctness alone) may allow models greater flexibility in learning attention patterns, while more structured objectives (i.e., requiring document citations or content reproduction) may provide stronger supervision but risk overfitting to specific output formats.
We define these reward functions for model output $y$ as follows, where $\text{ans}(y)$ extracts the answer, $\text{ids}(y)$ extracts cited document indices, $\text{cont}(y)$ extracts reproduced content, $\text{quotes}(y)$ extracts quoted spans, and $\approx$ denotes sub-exact match:

{\small
\begin{align}
R_{\text{AO}}(y) &= \mathds{1}[\text{ans}(y) \approx a^*] \label{eq:ao} \\
R_{\text{ID}}(y) &= \mathds{1}[\text{ids}(y) = \mathcal{D}^*]  + R_{\text{AO}}(y) \label{eq:ida} \\
R_{\text{ID+C}}(y) &= \mathds{1}[\text{cont}(y) \approx \{d_i : i \in \mathcal{D}^*\}] + R_{\text{ID}}(y) \label{eq:idca} \\
R_{\text{ID+Q}}(y) &= \mathds{1}[\text{quotes}(y) \subseteq \{d_i : i \in \mathcal{D}^*\}] + R_{\text{ID}}(y) \label{eq:idqa} \\
R_{\text{R+Judge}}(y) &= J(\text{reason}(y), \mathcal{D}_{\text{cited}}, \mathcal{D}^*, q, c) + R_{\text{AO}}(y) \label{eq:judge}
\end{align}
}

$R_{\text{AO}}(y)$ rewards solely correct answers. 
$R_{\text{ID}}(y)$ additionally encourages identifying all relevant document IDs before answering. 
$R_{\text{ID+C}}(y)$ further rewards the model if it reproduces the full text of relevant documents. 
$R_{\text{ID+Q}}(y)$ rewards extracting pertinent quotes from relevant documents rather than reproducing entire passages. 
$R_{\text{R+Judge}}(y)$ prompts the model to provide step-by-step reasoning, which an LLM judge $J$ evaluates for coherence and appropriate use of cited documents $\mathcal{D}_{\text{cited}}$.
Details about the reward functions can be found in Appendix~\ref{app:reward-functions}.

\section{Experimental Setup}\label{sec:exp}

\paragraph{Training.}\label{sec:exp-training-data}
For training data construction, we employ Qwen3-Embedding-4B~\cite{zhang2025qwen3embeddingadvancingtext} as the retriever and set $k=500$ to retrieve the top 500 passages for each question.
We then filter instances to retain only those with context lengths of at most 32k tokens.
This choice ensures two key advantages: \emph{i)} training remains computationally efficient; \emph{ii)} it allows us to evaluate whether models trained on these moderate-length contexts can generalize to much longer contexts, including 1M tokens.
Details about the prompts and hyperparameters used to train our models can be found in Appendix~\ref{app:training-hyper}.

\paragraph{Benchmarks.}\label{sec:exp-benchmarks}
We evaluate our models on a diverse set of benchmarks, domains and tasks.
For in-domain analysis, we select the HotpotQA~\cite{yang-etal-2018-hotpotqa}, NQ~\cite{kwiatkowski-etal-2019-natural} and TriviaQA~\cite{joshi-etal-2017-triviaqa} subsets from the RAG split of HELMET~\cite{yen2025helmetevaluatelongcontextlanguage}, where each instance contains a question with $p$ passages ($p-k$ distractors and $k$ relevant documents).
In these subsets, is it assumed that RAG has been applied to construct the input context; therefore, we do not perform additional retrieval.
For out-of-domain evaluation, we assess performance on $\infty$Bench~\cite{zhang-etal-2024-bench} (MC, QA, and SUM), LongBench-v2~\cite[LB-v2]{bai-etal-2025-longbench}, and the English financial split of Loong~\cite{wang-etal-2024-leave}, which requires reasoning over tables and numbers.
The prompts used in our experiments can be found in Appendix \ref{app:prompts}, while details on the evaluation metrics can be found in Appendix \ref{app:metrics}.

\paragraph{Models and Baselines.}\label{sec:exp-models}
Based on preliminary experiments comparing multiple LCLMs with context windows from 128k to 1M against their RAG counterparts (Appendix~\ref{app:model-selection}), we select Qwen2.5-7B-Instruct-1M~\cite{qwen2025qwen25technicalreport} as our base model. 
This model showed a clear performance gap between full-context and RAG, making it an ideal candidate for investigating whether fine-tuning can close this gap.
We use RAG@32k as our baseline, as it achieves the highest average performance.
Finally, to further motivate the choice of our RL-based approaches, we also use our dataset to train an SFT baseline on the answer-only objective, following~\cite{qiu-etal-2025-eliciting}.

\section{Results}\label{sec:results}

\begin{table}[t]
\centering
\footnotesize
\setlength{\tabcolsep}{4pt}
\begin{tabular}{@{}lcccc@{}}
\toprule
& \multicolumn{3}{c}{\textbf{HELMET (SubEM $\uparrow$)}} & \\
\cmidrule(lr){2-4}
\textbf{Model} & \textbf{HotpotQA} & \textbf{NQ} & \textbf{TriviaQA} & \textbf{Avg.} \\
\midrule
Base & 61.0 & 61.1 & 86.6 & 69.6 \\
\quad w/ gold docs & 72.2 & 70.7 & 85.8 & 76.2 \\
\midrule
+ $SFT_{\text{AO}}$ & 51.4 & 40.3 & 76.1 & 55.9 \\
\midrule
+ $R_{\text{AO}}(y)$ & \textbf{81.1} & 67.7 & 92.1 & \textbf{80.3} \\
+ $R_{\text{ID}}(y)$ & 75.2 & 65.5 & 91.0 & 77.2 \\
+ $R_{\text{ID+C}}(y)$ & 67.9 & 56.9 & 85.8 & 70.2 \\
+ $R_{\text{ID+Q}}(y)$ & 74.1 & 59.1 & 85.5 & 72.9 \\
+ $R_{\text{R+Judge}}(y)$ & 70.6 & \textbf{72.8} & \textbf{92.3} & 78.6 \\
\bottomrule
\end{tabular}
\caption{Average performance on HELMET's ``RAG'' subset across context lengths 4k--128k.}
\label{tab:in-domain-results}
\end{table}

\begin{table}[t]
\centering
\footnotesize
\setlength{\tabcolsep}{2.5pt}
\begin{tabular}{@{}lcccccc@{}}
\toprule
& \multicolumn{3}{c}{\textbf{$\infty$Bench}} & \textbf{LB-v2} & \textbf{Loong} & \\
\cmidrule(lr){2-4} \cmidrule(lr){5-5} \cmidrule(lr){6-6}
\textbf{Model} & \textbf{MC} & \textbf{QA} & \textbf{Sum} & \textbf{ALL} & \textbf{Fin.} & \textbf{Avg.} \\
& \scriptsize{(Acc. $\uparrow$)} & \scriptsize{(SubEM $\uparrow$)} & \scriptsize{(R-L $\uparrow$)} & \scriptsize{(Acc. $\uparrow$)} & \scriptsize{(Judge $\uparrow$)} & \\
\midrule
\multicolumn{7}{@{}l}{\textit{Full Context (up to 1M tokens)}} \\
\midrule
Base & 72.0 & 30.0 & 31.2 & 32.0 & 49.9 & 43.0 \\
+ $R_{\text{AO}}(y)$ & 70.3 & \textbf{38.2} & 31.0 & 31.2 & 55.3 & 45.2 \\
+ $R_{\text{ID}}(y)$ & 71.6 & 34.7 & 29.7 & 31.0 & 51.2 & 43.6 \\
+ $R_{\text{ID+C}}(y)$ & 70.7 & 31.3 & 30.2 & 31.4 & 47.5 & 42.2 \\
+ $R_{\text{ID+Q}}(y)$ & 70.7 & 30.8 & 30.5 & 32.0 & 51.5 & 43.1 \\
+ $R_{\text{R+Judge}}(y)$ & 72.5 & 30.7 & \textbf{31.7} & 32.6 & \textbf{58.8} & \textbf{45.3} \\
\midrule
\multicolumn{7}{@{}l}{\textit{KV-Cache with RetrievalAttn}} \\
\midrule
Base & 62.0 & 27.6 & 27.3 & 30.2 & 39.5 & 37.3 \\
+ $R_{\text{AO}}(y)$ & 65.5 & 33.6 & 28.0 & 29.6 & 41.7 & 39.7 \\
+ $R_{\text{ID}}(y)$ & 64.2 & 29.3 & 26.0 & 30.0 & 43.0 & 38.5 \\
+ $R_{\text{ID+C}}(y)$ & 63.3 & 27.1 & 27.7 & 29.6 & 39.8 & 37.5 \\
+ $R_{\text{ID+Q}}(y)$ & 62.9 & 27.1 & 28.1 & 32.0 & 39.7 & 38.0 \\
+ $R_{\text{R+Judge}}(y)$ & 63.8 & 28.2 & 28.8 & 31.2 & 41.5 & 38.7 \\
\midrule
\multicolumn{7}{@{}l}{\textit{RAG (Base Model)}} \\
\midrule
RAG@32k & \textbf{78.6} & 29.3 & 25.1 & \textbf{38.0} & 47.8 & 43.8 \\
\bottomrule
\end{tabular}
\caption{Out-of-domain performance comparison.}
\label{tab:ood-results}
\end{table}

\begin{table}[t]
\centering
\small
\setlength{\tabcolsep}{0pt}
\begin{tabular*}{\columnwidth}{@{\extracolsep{\fill}}lrrrrrc@{}}
\toprule
& \multicolumn{6}{c}{\textbf{Performance Drop (\%)}} \\
\cmidrule(lr){2-7}
\textbf{Model} & \textbf{MC} & \textbf{QA} & \textbf{Sum} & \textbf{ALL} & \textbf{Fin} & \textbf{Avg} \\
\midrule
Base & -13.9 & -27.6 & -12.5 & -5.6 & -20.8 & -16.1 \\
+ $R_{\text{AO}}(y)$ & \textbf{-6.8} & -28.3 & -9.7 & -5.1 & -24.6 & -14.9 \\
+ $R_{\text{ID}}(y)$ & -10.3 & -28.3 & -12.5 & -3.2 & \textbf{-16.0} & -14.1 \\
+ $R_{\text{ID+C}}(y)$ & -10.5 & -28.7 & -8.3 & -5.7 & -16.2 & \textbf{-13.9} \\
+ $R_{\text{ID+Q}}(y)$ & -11.0 & -32.7 & \textbf{-7.9} & \textbf{0.0} & -22.9 & -14.9 \\
+ $R_{\text{R+Judge}}(y)$ & -12.0 & -30.2 & -10.5 & -4.5 & -18.7 & -15.2 \\
\bottomrule
\end{tabular*}
\caption{Performance drop (\%) with RA vs. full context. MC/QA/Sum: $\infty$Bench, ALL: LB-v2, Fin: Loong.}
\label{tab:retrieval-attention-drop}
\end{table}

\begin{table}[t]
\centering
\small
\begin{tabular}{@{}lccccc@{}}
\toprule
\multirow{2}{*}{\textbf{Model}} & \multicolumn{3}{c}{\textbf{Documents Ranking (NDCG@10)}} & \\
\cmidrule(lr){2-4}
& \textbf{HotpotQA} & \textbf{NQ} & \textbf{TriviaQA} & \textbf{Avg.} \\
\midrule
Base & 82.0 & 84.7 & 83.6 & 83.4 \\
+ $R_{\text{AO}}(y)$ & 82.3 & 85.0 & 83.8 & 83.7 \\
+ $R_{\text{ID}}(y)$ & 82.3 & 85.0 & 83.7 & 83.7 \\
+ $R_{\text{ID+C}}(y)$ & 82.4 & 85.1 & 84.1 & 83.9 \\
+ $R_{\text{ID+Q}}(y)$ & \textbf{82.5} & \textbf{85.4} & \textbf{84.3} & \textbf{84.1} \\
+ $R_{\text{R+Judge}}(y)$ & 82.2 & 84.9 & 83.7 & 83.6 \\
\bottomrule
\end{tabular}
\caption{NDCG@10 computed on reranked documents on HELMET at 32k context length.}
\label{tab:attention-scores}
\end{table}

\paragraph{In-Domain Results.}\label{sec:results-id}

Table~\ref{tab:in-domain-results} presents average performance on the HELMET's RAG subset across context lengths from 4k to 128k tokens (see Appendix~\ref{app:detailed-results} for per-length results). 
From the results we can immediately see how standard SFT leads to severe performance degradation compared to the base model, resulting in an average drop of 13.7 points (55.9 vs. 69.6), while RL-based approaches consistently outperform the baseline.
The $R_{\text{AO}}(y)$ reward function achieves the strongest improvements on HotpotQA, with gains of 20.1 points (81.1 vs. 61.0) over the base model, while $R_{\text{R+Judge}}(y)$ achieves the best performance on NQ (72.8 vs. 61.1) and TriviaQA (92.3 vs. 86.6).
The $R_{\text{ID}}(y)$ reward function also yields substantial gains, averaging 75.2 on HotpotQA, 65.5 on NQ, and 91.0 on TriviaQA. 
However, more constrained strategies, specifically $R_{\text{ID+C}}(y)$ and $R_{\text{ID+Q}}(y)$, show performance degradation, suggesting that overly restrictive training objectives may impede the model's ability to process extended contexts effectively.
Notably, both $R_{\text{AO}}(y)$ and $R_{\text{R+Judge}}(y)$ surpass the performance obtained when the base model is informed about which documents are relevant: $R_{\text{AO}}(y)$ exceeds gold docs on HotpotQA (81.1 vs. 72.2) and TriviaQA (92.1 vs. 85.8), while $R_{\text{R+Judge}}(y)$ surpasses gold docs on NQ (72.8 vs. 70.7) and TriviaQA (92.3 vs. 85.8).
This indicates that fine-tuning successfully enables models to identify relevant information without sacrificing the model's ability to leverage its full context.

\paragraph{Out-of-Domain Results.}\label{sec:results-ood}

Table~\ref{tab:ood-results} evaluates generalization to out-of-domain benchmarks where inputs consist of long, cohesive documents rather than retrieved passages. 
Results reveal mixed patterns across tasks and models.
$R_{\text{R+Judge}}(y)$ demonstrates the most substantial improvement on Loong Financial, achieving 58.8 (+8.9 over base) and outperforming RAG@32k by 11 points (58.8 vs. 47.8).
Similarly, $R_{\text{AO}}(y)$ yields the largest gain on $\infty$Bench QA (38.2, +8.2 points), also surpassing RAG@32k (38.2 vs. 29.3).
However, on other benchmarks (LB-v2, $\infty$Bench MC and SUM) we do not observe substantial improvement: multiple-choice accuracy remains largely unchanged (70.3-72.5 vs. 72.0 base), and summarization shows minimal variation (30.2-31.7 vs. 31.2 base).
Conversely, RAG at 32k tokens outperforms all fine-tuned models on $\infty$Bench MC (78.6\% vs. 72.5\% best fine-tuned) and LB-v2 (38.0\% vs. 32.6\% best fine-tuned).
We attribute this variance to our training objective, which explicitly optimizes for sparse retrieval by focusing on identifying specific, isolated facts. This specialization may limit performance on tasks requiring dense information integration over cohesive documents, explaining why standard RAG pipelines retain an advantage in these specific scenarios.
These mixed results indicate that no single approach consistently dominates across diverse out-of-domain tasks, with both fine-tuning and RAG offering complementary strengths.

\paragraph{KV-Cache Compression Analysis.}\label{sec:results-kv}

To assess whether fine-tuning translates to stable performance in efficient inference settings, we evaluate models under RetrievalAttention~\cite[RA]{liu2025retrievalattention}, a KV-cache compression technique that retains only the top-k most-attended key-value pairs.
We select RA because it directly aligns with our training objective: models that allocate more attention to relevant content should theoretically retain performance when compression preserves only highly-attended tokens.

Tables~\ref{tab:ood-results} and~\ref{tab:retrieval-attention-drop} quantify performance under RA, showing that all models experience substantial degradation, with the base model dropping 16.1\% on average.
Fine-tuned models reduce this degradation moderately: $R_{\text{ID+C}}(y)$ achieves the smallest drop (13.9\%), representing 2.2 percentage point improvements over the base model.
However, even the best compressed fine-tuned models fall short of RAG@32k's performance, indicating that compression still incurs performance costs that fine-tuning only partially mitigates.
Robustness patterns vary considerably by task: $R_{\text{AO}}(y)$ shows the best resilience on $\infty$Bench MC (6.8\% drop vs. 13.9\% base) but the worst on Loong Financial (24.6\% drop vs. 20.8\% base), while $R_{\text{ID+Q}}(y)$ maintains perfect performance on LB-v2 (0.0\% drop) yet degrades substantially on $\infty$Bench QA (32.7\% drop).
Our findings demonstrate that while fine-tuning can improve task performance and reduce degradation under compression, current approaches yield moderate gains that remain highly task-specific, highlighting the need for more targeted optimization strategies for efficient long-context inference.

\paragraph{Document Ranking Analysis.}\label{sec:results-attn}

To investigate whether performance improvements stem from enhanced selective attention, we measure document ranking quality using NDCG@10 at 32k context length (our training setup), ranking documents by cumulative attention scores and comparing against HELMET's relevance-based ordering.

Table~\ref{tab:attention-scores} shows that the base model already achieves high NDCG@10 scores (82.0--84.7).
Fine-tuning yields minimal improvements: $R_{\text{ID+Q}}(y)$ achieves the highest scores with only 0.5--0.7 point gains over the base.
However, the Pearson correlation between attention ranking and task performance is negligible (r=-0.09, p=0.86): $R_{\text{ID+Q}}(y)$ achieves the best NDCG@10 but poor accuracy (72.9), while $R_{\text{AO}}(y)$ and $R_{\text{R+Judge}}(y)$ demonstrate strong performance (80.3, 78.6) despite comparable NDCG@10 scores.
These findings suggest that improved attention-based document ranking does not explain the observed performance gains. 
Instead, since the base model already successfully reranks relevant information, we attribute the in- and out-of-domain improvements to enhanced task robustness. 
The GRPO-trained models demonstrate superior discrimination in noisy contexts, allowing them to effectively utilize the attended information and filter out distractions where the base model, despite similar attention patterns, fails.

\section{Conclusions}\label{sec:conclusions}
In this work, we investigated whether fine-tuning LCLMs to selectively attend to relevant information allows them to effectively replace conventional RAG systems. 
We found that while standard SFT often degrades capabilities, our proposed GRPO-based strategies, particularly those leveraging answer-only and reasoning rewards, yield substantial in-domain improvements, successfully bridging the performance gap with RAG. 
However, out-of-domain generalization remains inconsistent. 
Our results suggest that while our models excel at sparse retrieval tasks involving specific fact extraction, they struggle with tasks requiring dense information integration, where RAG retains an advantage.

Regarding efficiency, we demonstrated that fine-tuning mitigates performance degradation under KV-cache compression, although these gains are moderate and task-dependent. 
Crucially, our analysis indicates that performance improvements do not stem from better attention-based document ranking, which correlates weakly with accuracy. 
Instead, the gains appear driven by enhanced robustness to noise within the context. 
We conclude that while LCLMs offer a powerful alternative to RAG, achieving universal generalization requires future training objectives that better balance selective attention with the ability to synthesize dense information.

\section*{Acknowledgments}
We thank Ionut-Teodor Sorodoc, Gianni Barlacchi, Dennis Fucci, Nathanaël Carraz Rakotonirina, and Sebastian Steindl for their insightful discussions and valuable feedback throughout the project. We are also grateful to the anonymous reviewers for their constructive comments, which helped improve this paper.

Bill Byrne holds concurrent appointments as an Amazon Scholar and as Professor of Information Engineering at the University of Cambridge. This paper describes work performed at Amazon.

\section*{Limitations}

We train our models on contexts up to 32k tokens both for computational efficiency and for evaluating generalization on longer inputs, up to 1M tokens. 
This design choice allows us to assess generalization to longer contexts but limits our ability to investigate whether training on extremely long sequences would yield different outcomes.
Given sufficient computational resources, it would be valuable to extend training to 1M context lengths, though our observed generalization patterns suggest findings would remain largely consistent.

Moreover, our evaluation focuses on English-language benchmarks in question answering, multiple choice, summarization, and financial analysis domains. 
The effectiveness of these fine-tuning strategies for other languages, modalities, or task types remains an open question.

Finally, our training data is constructed from Wikipedia passages due to the availability of gold-standard annotations in HotpotQA and 2WikiMultihopQA.
While this enables controlled experimentation with verifiable rewards, the lack of domain-specific annotated data (e.g.,~legal documents, scientific papers, technical manuals) limits our ability to assess whether domain-adapted training would improve out-of-domain generalization.
Given sufficient annotated data containing gold documents for other domains, our approach could be extended to enable more robust cross-domain performance.

\bibliography{ms}

\clearpage

\appendix

\section{Training Data Details}\label{app:training-data}

\paragraph{Hard Negative Refinement.}
Since retrieving hard negatives from Wikipedia can inadvertently include documents that are actually relevant to the question, we apply a two-stage preprocessing pipeline to identify and promote such documents from hard negatives to relevant passages.
In the first stage, we apply fuzzy matching heuristics, including Jaccard similarity, character-level F1, and n-gram matching, between each gold passage and each hard negative, using a loose similarity threshold.
This ensures high recall, capturing the vast majority of true positives within the input context.
However, this approach can introduce false positives; therefore, in the second stage, we employ a strong LLM as a judge to filter out promoted documents that are indeed false positives.
This two-stage procedure results in a refined set of truly relevant documents, increasing the average number of relevant passages from 2 to approximately 4 per instance.
This refinement makes the model more resilient to out-of-domain cases where more than 2 passages may be needed to answer the question.

In our experiments, we use a fuzzy similarity threshold of 0.6 for the initial matching stage.
For the second stage, we employ DeepSeek-R1-Distill-Qwen-32B~\cite{DeepSeekAI2025DeepSeekR1IR} with thinking enabled and temperature set to 0.0 as the judge to filter false positives.

\paragraph{Training Data Examples.}
Table~\ref{tab:training-examples} shows representative instances from our training data, highlighting the challenging nature of the task.
Each example contains a question paired with a long context comprising multiple documents, where only a small subset (indicated by Gold IDs and shown in bold) contains information relevant to answering the question.
The majority of documents serve as hard negatives, which are topically related but irrelevant passages that create a realistic information retrieval scenario.
For instance, in the first example, while many documents discuss various magazines, only documents 10 and 11 contain the specific founding dates needed to determine which magazine was started first.
Similarly, the second example includes multiple documents about Indian hotel chains and cities, but only documents 0 and 4 provide the critical information linking the Oberoi family to Delhi.
We posit that this sparse relevant information embedded within extensive hard negatives may force models to develop selective attention mechanisms, mimicking the challenges faced by RAG systems when processing retrieved contexts.

\begin{table*}[t]
\centering
\footnotesize
\setlength{\tabcolsep}{4pt}
\begin{tabular}{@{}p{0.28\textwidth}|p{0.12\textwidth}|p{0.48\textwidth}|p{0.08\textwidth}@{}}
\toprule
\textbf{Question} & \textbf{Answer} & \textbf{Training Context (abbreviated)} & \textbf{Gold IDs} \\
\midrule
Which magazine was started first Arthur's Magazine or First for Women? 
& Arthur's Magazine 
& [DOC 0] British art critics ... [DOC 1] The English Woman's Journal ... [DOC 2] 1300 subscribers ... [DOC 3] Magazines published in UK ... [DOC 4] French male writers ... [DOC 5] Published in USA ... [DOC 6] Magazines with year ... [DOC 7] Feminism and family ... [DOC 8] UK Monthly magazines ... [DOC 9] (magazine) ... \textbf{[DOC 10] Arthur's Magazine (1844–1846) American literary periodical ...} \textbf{[DOC 11] First for Women ... woman's magazine published by Bauer Media ...}
& 10, 11 \\
\midrule
The Oberoi family is part of a hotel company that has a head office in what city?
& Delhi
& \textbf{[DOC 0] The Oberoi Group is a hotel company with its head office in Delhi.} [DOC 1] Taj Hotels ... headquartered in Mumbai ... [DOC 2] Natural History Society ... [DOC 3] Taj Dubai ...  \textbf{[DOC 4] The Oberoi family is an Indian family involved in hotels through The Oberoi Group.} [DOC 5] Food and drink companies based in Boston ... [DOC 6] City serves as headquarters ... Kolkata ... [DOC 7] Bangalore, Taj Connemara ... [DOC 8] Observer Research Foundation based in Delhi ... [DOC 9] Companies based in Miami ... [DOC 10] Survey of India ... Kolkata ...
& 0, 4 \\
\midrule
Musician and satirist Allie Goertz wrote a song about the "The Simpsons" character Milhouse, who Matt Groening named after who?
& Richard Nixon
& [DOC 0] Simpson ... Harry Shearer as Mr. Burns ... [DOC 1] Jon Lovitz as Artie Ziff ... \textbf{[DOC 2] Allison Beth "Allie" Goertz (born March 2, 1991) American musician.} [DOC 3] Krusty the Clown ... Bart ... \textbf{[DOC 4] Her videos are posted on YouTube under Cossbysweater.} [DOC 5] Simpsons ... Jerry Nelson ... \textbf{[DOC 6] Goertz is known for her satirical songs based on pop culture topics.} [DOC 7] Ralph Wiggum ... \textbf{[DOC 8] Milhouse Mussolini van Houten ... fictional character ... created by Matt Groening ...} [DOC 9] Guitar ... parody ... [DOC 10] by "Weird Al" Yankovic ... [DOC 11] Beverly Hills 90210 ... [DOC 12] Baby bunnies ... [DOC 13] Allegra's Window ...
& 2, 4, 6, 8 \\
\bottomrule
\end{tabular}
\caption{Examples of training data instances showing questions with long-context inputs containing both relevant documents (bold) and hard negative distractors. Gold IDs indicate the document indices required to answer each question.}
\label{tab:training-examples}
\end{table*}

\section{Reward Functions}\label{app:reward-functions}
We define the following reward functions for model output $y$:

\paragraph{Answer Only (AO).}
The model receives a reward for generating the correct answer:
\vspace{-1mm}
\begin{equation}
R_{\text{AO}}(y) = \mathds{1}[\text{ans}(y) \approx a^*]
\end{equation}
\vspace{-1mm}
\noindent where $\text{ans}(y)$ extracts the answer from output $y$, and $\approx$ denotes sub-exact match allowing for minor lexical variations.
\paragraph{Document IDs + Answer (ID).}
The model is rewarded for correctly identifying relevant document IDs and providing the correct answer:
\begin{equation}
R_{\text{ID}}(y) = \mathds{1}[\text{ids}(y) = \mathcal{D}^*] + \mathds{1}[\text{ans}(y) \approx a^*]
\end{equation}
where $\text{ids}(y)$ extracts the cited document indices from the output.

\paragraph{Document IDs + Content + Answer (ID+C).}
The model is rewarded for generating the IDs and reproducing the content of relevant documents:
\begin{equation}
\begin{split}
R_{\text{ID+C}}(y) = \mathds{1}[\text{ids}(y) = \mathcal{D}^*] + \\
\mathds{1}[\text{cont}(y) \approx \{d_i : i \in \mathcal{D}^*\}] + \mathds{1}[\text{ans}(y) \approx a^*]
\end{split}
\end{equation}
where $\text{cont}(y)$ extracts the document content reproduced in the output.

\paragraph{Document IDs + Quotes + Answer (ID+Q).}
The model is rewarded for identifying relevant documents and extracting pertinent quotes:
\begin{equation}
\begin{split}
R_{\text{ID+Q}}(y) = \mathds{1}[\text{ids}(y) = \mathcal{D}^*] + \\
\mathds{1}[\text{quotes}(y) \subseteq \{d_i : i \in \mathcal{D}^*\}] + \mathds{1}[\text{ans}(y) \approx a^*]
\end{split}
\end{equation}
where $\text{quotes}(y)$ extracts quoted text spans from the output.

\paragraph{Reasoning + LLM-as-a-Judge (R+Judge).}
The policy is prompted to reason over the context and justify its answer.
We parse the output to extract cited document IDs $\mathcal{D}_{\text{cited}}$ and employ an LLM judge $J$ to evaluate relevance:
\begin{equation}
\begin{split}
R_{\text{R+Judge}}(y) = J(\text{reason}(y), \mathcal{D}_{\text{cited}}, \mathcal{D}^*, q, c) + \\
\mathds{1}[\text{ans}(y) \approx a^*]
\end{split}
\end{equation}
The judge $J$ assigns a score based on whether cited documents are in $\mathcal{D}^*$ or contain information genuinely useful for answering the question.
This relaxed reward addresses the limitation that annotated relevant documents may be incomplete, as topically related documents can provide valid reasoning paths.

\section{Training Settings}\label{app:training-hyper}

\begin{table*}[t]
\centering
\footnotesize
\begin{tabular}{@{}p{0.96\textwidth}@{}}
\toprule
\textbf{LLM Judge Evaluation Prompt Template} \\
\midrule
\texttt{You are an expert evaluator assessing AI model answers to questions using supporting documents.} \\
\texttt{You will be provided with:} \\
\texttt{- A \textbf{Question}} \\
\texttt{- A set of \textbf{Relevant Documents} (the gold standard grounding sources)} \\
\texttt{- The \textbf{Correct Answer}} \\
\texttt{- An \textbf{AI Model Solution}} \\
\\
\texttt{\textbf{Background:}} \\
\texttt{The AI model had access to a large pool of documents (indexed 0...N). Only a subset is truly} \\
\texttt{relevant. Other documents may appear in citations but are simply distractors (not fabricated).} \\
\texttt{The model's goal is to correctly answer the question while grounding its reasoning in the} \\
\texttt{relevant documents.} \\
\\
\texttt{\textbf{Your task:}} \\
\texttt{Evaluate the model's solution objectively and consistently according to the criteria below.} \\
\texttt{Do not use information outside the provided inputs.} \\
\\
\texttt{---} \\
\\
\texttt{[Question]: \{question\}} \\
\texttt{[Relevant Documents]: \{gold\_docs\}} \\
\texttt{[Correct Answer]: \{answer\}} \\
\texttt{[AI Model Solution]: \{solution\}} \\
\\
\texttt{---} \\
\\
\texttt{\textbf{EVALUATION CRITERIA}} \\
\\
\texttt{\textbf{Criterion 1: Reasoning Quality (1 or 0)}} \\
\texttt{Score \textbf{1} if the solution shows: clear logical flow from evidence to conclusion, no} \\
\texttt{contradictions or fallacies, and coherent, well-structured reasoning.} \\
\texttt{Score \textbf{0} if the reasoning is flawed, contradictory, or incoherent.} \\
\\
\texttt{\textbf{Criterion 2: Document Grounding (1 or 0)}} \\
\texttt{Score \textbf{1} if the solution: uses information primarily from relevant documents, represents} \\
\texttt{those documents accurately (no distortions), and does not rely significantly on irrelevant or} \\
\texttt{external knowledge. Score \textbf{0} if it misuses documents or ignores relevant evidence.} \\
\\
\texttt{\textbf{Criterion 3: Answer Correctness (1 or 0)}} \\
\texttt{Score \textbf{1} if the final answer matches the provided correct answer.} \\
\texttt{Score \textbf{0} otherwise (including partial or incomplete answers).} \\
\\
\texttt{---} \\
\\
\texttt{\textbf{RESPONSE FORMAT}} \\
\texttt{For each criterion, provide a 1-2 sentence justification followed by the score:} \\
\\
\texttt{Reasoning Quality Justification: [Your explanation]} \\
\texttt{$\backslash$boxed\{Criterion 1: 1 or 0\}} \\
\\
\texttt{Document Grounding Justification: [Your explanation]} \\
\texttt{$\backslash$boxed\{Criterion 2: 1 or 0\}} \\
\\
\texttt{Answer Correctness Justification: [Your explanation]} \\
\texttt{$\backslash$boxed\{Criterion 3: 1 or 0\}} \\
\bottomrule
\end{tabular}
\caption{Prompt template for LLM judge evaluation in the R+Judge reward function. The judge evaluates model outputs across three binary criteria: reasoning quality, document grounding, and answer correctness. The final reward is computed as the sum of all three scores.}
\label{tab:training-judge-prompt}
\end{table*}

We train all models using GRPO with the VERL framework~\cite{sheng2024hybridflow} and FSDP for distributed training.
We split the data into 95\% train and 5\% development, resulting in 28,500 training samples and 1,500 validation samples.
The training batch size is set to 256 and validation batch size to 128.
We set the maximum prompt length to 32,768 tokens to accommodate long contexts, with a maximum response length of 2,048 tokens, applying middle truncation for sequences exceeding these limits.
We use the AdamW optimizer with a learning rate of $1 \times 10^{-6}$, a PPO mini-batch size of 64, and a micro-batch size of 2 per GPU.
Training proceeds for up to 2 epochs with model checkpointing every 10 steps.
For GRPO, we sample $G=5$ outputs from the current policy for each training instance to compute group-relative advantages.
The KL divergence coefficient $\beta$ is set to 0.001 with a low-variance KL loss formulation, applied only in the loss function and not in the reward computation itself.
To optimize memory usage, we enable gradient checkpointing and activation offloading, with Ulysses sequence parallelism~\cite{jacobs2023deepspeed} set to 4 and tensor model parallelism of 2 for rollout generation.
We use vLLM for efficient rollout generation with chunked prefill enabled and a maximum of 34,816 batched tokens, while reference model computations use FSDP with parameter offloading. We cap GPU memory utilization at 60\%.
In Table \ref{tab:training-judge-prompt} we provide the judge prompt used to train our models with the reward function $R_{\text{R+Judge}}(y)$ (see Section \ref{sec:methodology-training} for details).

\section{Evaluation Prompts}\label{app:prompts}
Table~\ref{tab:eval-prompts} presents the prompt templates used for evaluation across all out-of-domain benchmarks.
All prompts are taken directly from the official benchmark papers.
The \texttt{\{instruction\}} component in the Loong financial split can be found in the original paper~\cite{wang-etal-2024-leave}.
For the in-domain benchmark HELMET, each model trained with a different reward function is evaluated using a distinct prompt structure that matches its training objective.
Tables~\ref{tab:helmet-prompts-1} and~\ref{tab:helmet-prompts-2} present the complete prompt templates for all five reward functions applied to the RAG split of HELMET.
These prompts are designed to elicit the specific output formats required for computing rewards during training, ensuring consistency between training and evaluation.

\begin{table*}[t]
\centering
\footnotesize
\begin{tabular}{@{}p{0.18\textwidth}p{0.78\textwidth}@{}}
\toprule
\textbf{Benchmark} & \textbf{Prompt Template} \\
\midrule
\multicolumn{2}{@{}l}{\textit{$\infty$Bench}} \\
\midrule
QA Split & 
\texttt{System: You are an expert in answering questions about books.} \\
& \texttt{User: Read the book below and answer the question.} \\
& \\
& \texttt{\{context\}} \\
& \\
& \texttt{Question: \{question\}} \\
& \\
& \texttt{Format your response as follows: "The answer is (insert answer here)".} \\
\midrule
SUM Split & 
\texttt{System: You are an expert in summarizing books.} \\
& \texttt{User: Summarize the book below.} \\
& \\
& \texttt{\{context\}.} \\
\midrule
MC Split & 
\texttt{System: You are an expert in multiple-choice question answering about books.} \\
& \texttt{User: Read the book below and answer the question.} \\
& \\
& \texttt{\{context\}} \\
& \\
& \texttt{Question: \{question\}} \\
& \\
& \texttt{Only one of the following options is correct:} \\
& \\
& \texttt{A. \{choice\_a\}} \\
& \texttt{B. \{choice\_b\}} \\
& \texttt{C. \{choice\_c\}} \\
& \texttt{D. \{choice\_d\}} \\
& \\
& \texttt{Format your response as follows: "The answer is (insert answer here)".} \\
\midrule
\multicolumn{2}{@{}l}{\textit{LB-v2 (ALL split)}} \\
\midrule
ALL Split & 
\texttt{System: You are an expert in multiple-choice question answering.} \\
& \texttt{User: Please read the following text and answer the question below.} \\
& \\
& \texttt{<text>} \\
& \texttt{\{DOC\}} \\
& \texttt{</text>} \\
& \\
& \texttt{What is the correct answer to this question: \{Q\}} \\
& \texttt{Choices:} \\
& \texttt{(A) \{C\_A\}} \\
& \texttt{(B) \{C\_B\}} \\
& \texttt{(C) \{C\_C\}} \\
& \texttt{(D) \{C\_D\}} \\
& \\
& \texttt{Format your response as follows: "The correct answer is (insert answer here)".} \\
\midrule
\multicolumn{2}{@{}l}{\textit{Loong}} \\
\midrule
Financial Split & 
\texttt{System: You are an expert in answering questions about financial reports.} \\
& \texttt{User: \{context\}} \\
& \\
& \texttt{\{instruction\}} \\
& \\
& \texttt{\{question\}.} \\
\bottomrule
\end{tabular}
\caption{Prompt templates used for evaluation across out-of-domain benchmarks. All prompts follow a consistent format with system instructions defining the task domain and user prompts providing context and questions. Variables enclosed in braces (e.g.,~\texttt{\{context\}}, \texttt{\{question\}}) are replaced with actual benchmark data during evaluation.}
\label{tab:eval-prompts}
\end{table*}

\begin{table*}[t]
\centering
\footnotesize
\begin{tabular}{@{}p{0.20\textwidth}p{0.76\textwidth}@{}}
\toprule
\textbf{Reward Function} & \textbf{Prompt Template (KILT split)} \\
\midrule
Answer Only (AO) & 
\texttt{System: You are an expert in question answering.} \\
& \texttt{User: Use the given documents to write a concise and short answer to the question.} \\
& \\
& \texttt{\{context\}} \\
& \\
& \texttt{Question: \{question\}} \\
& \\
& \texttt{Write your concise and short answer in the following format:} \\
& \\
& \texttt{Answer: <your answer>.} \\
\midrule
ID + Answer (ID) & 
\texttt{System: You are an expert in question answering.} \\
& \texttt{User: Documents: \{context\}} \\
& \\
& \texttt{Instruction: Given the set of documents above, answer the following question by following these steps:} \\
& \\
& \texttt{1. Identify the relevant documents. Each document is identified by a tag in the form [DOC i], where i is the index of the document in the context.} \\
& \texttt{   - If you find relevant documents, output their IDs exactly as shown, separated by commas. Example: [DOC i], [DOC j], [DOC k].} \\
& \texttt{   - If no documents are relevant, output only: [DOC -1].} \\
& \\
& \texttt{2. On a new line, provide your answer in the following format:} \\
& \texttt{   The answer is: <your answer here>.} \\
& \\
& \texttt{Question: \{question\}.} \\
\midrule
ID + Content + Answer (ID+C) & 
\texttt{System: You are an expert in question answering.} \\
& \texttt{User: Documents: \{context\}} \\
& \\
& \texttt{Instruction: You are given a set of documents and a question. Your task is to:} \\
& \\
& \texttt{1. Identify which documents are most relevant to answering the question} \\
& \texttt{2. Extract and reproduce the IDs and the full content of those relevant documents} \\
& \texttt{3. Provide the final answer based on the relevant documents} \\
& \\
& \texttt{Follow this exact format in your response:} \\
& \\
& \texttt{Relevant documents:} \\
& \texttt{[DOC X]} \\
& \texttt{<full content of first relevant document>} \\
& \\
& \texttt{[DOC Y]} \\
& \texttt{<full content of second relevant document>} \\
& \\
& \texttt{(continue for all relevant documents)} \\
& \\
& \texttt{The answer is: <your final answer>} \\
& \\
& \texttt{Important guidelines:} \\
& \texttt{- Only include documents that directly help answer the question} \\
& \texttt{- Reproduce the ID and the complete content of each relevant document exactly} \\
& \texttt{- The final answer should be concise and directly address the question} \\
& \\
& \texttt{Question: \{question\}.} \\
\bottomrule
\end{tabular}
\caption{HELMET prompt templates: Part 1.}
\label{tab:helmet-prompts-1}
\end{table*}

\begin{table*}[t]
\centering
\footnotesize
\begin{tabular}{@{}p{0.20\textwidth}p{0.76\textwidth}@{}}
\toprule
\textbf{Reward Function} & \textbf{Prompt Template (KILT split)} \\
\midrule
ID + Quote + Answer (ID+Q) & 
\texttt{System: You are an expert in question answering.} \\
& \texttt{User: Documents: \{context\}} \\
& \\
& \texttt{Instruction: Given the set of documents above, answer the following question by following these steps:} \\
& \\
& \texttt{1. Extract relevant quotes: Find and extract short quotes or passages ($\leq$30 tokens each) from the documents that help answer the question. Present each quote in the following format:} \\
& \texttt{   Quote 1: "<exact text from document>"} \\
& \texttt{   Quote 2: "<exact text from document>"} \\
& \texttt{   (Continue as needed)} \\
& \\
& \texttt{2. List the source documents using this exact format:} \\
& \texttt{   Relevant Document IDs: [DOC i], [DOC j], [DOC k]} \\
& \texttt{   (Where i, j, k are the indices of documents that contain your selected quotes)} \\
& \texttt{   If no documents are relevant, use: Relevant Document IDs: [DOC -1]} \\
& \\
& \texttt{3. Provide your final answer in the following format:} \\
& \texttt{   The answer is: <your answer here>} \\
& \\
& \texttt{Important: Keep quotes short ($\leq$30 tokens), select only the most relevant passages, and ensure your document IDs correspond to the documents containing your quotes.} \\
& \\
& \texttt{Question: \{question\}.} \\
\midrule
Reasoning + Judge (R+Judge) & 
\texttt{System: You are an expert in question answering.} \\
& \texttt{User: Documents: \{context\}} \\
& \\
& \texttt{Instruction: You are given a set of documents and a question. Your task is to analyze all documents, reason through the question step-by-step, and provide a well-grounded answer.} \\
& \\
& \texttt{Follow this exact format in your response:} \\
& \\
& \texttt{**Step 1: Question Analysis**} \\
& \texttt{Break down what the question is asking and identify the key information needed.} \\
& \\
& \texttt{**Step 2: Document Review**} \\
& \texttt{Briefly assess which documents contain relevant information for answering the} \\
& \texttt{question. Consider how each document relates to the question's requirements.} \\
& \\
& \texttt{**Step 3: Reasoning**} \\
& \texttt{Work through the problem step-by-step using information from the relevant documents. For each reasoning step, clearly reference which document(s) support your logic. Use the format: 'According to [DOC X], ...' or 'Document Y states that ...' to ground your reasoning in the provided sources.} \\
& \\
& \texttt{**Step 4: Answer**} \\
& \texttt{The answer is: <your final answer>} \\
& \\
& \texttt{Important guidelines:} \\
& \texttt{- Show your complete reasoning process} \\
& \texttt{- Ground each reasoning step in specific document evidence} \\
& \texttt{- Clearly reference document IDs when using information from them} \\
& \texttt{- Keep your final answer concise but ensure it directly addresses the question} \\
& \\
& \texttt{Question: \{question\}} \\
\bottomrule
\end{tabular}
\caption{HELMET prompt templates: Part 2.}
\label{tab:helmet-prompts-2}
\end{table*}

\section{Metrics}\label{app:metrics}
We employ benchmark-specific evaluation metrics tailored to each task type.
For multiple-choice question answering benchmarks (LB-v2 and $\infty$Bench MC), we use accuracy as the primary metric.
To extract model answers, we employ xFinder~\cite{yu2025xfinderlargelanguagemodels}, an LLM-based answer extraction tool that has demonstrated higher agreement with human judgment compared to standard regex-based extractors~\cite{molfese-etal-2025-right}.
For open-ended question answering tasks ($\infty$Bench QA and HELMET), we use sub-exact match following the recommendation of \citet{yen2025helmetevaluatelongcontextlanguage}, who demonstrate that this metric better correlates with true model performance than strict exact match.
Sub-exact match allows for minor lexical variations while still requiring the core answer to be correct.
For the summarization task in $\infty$Bench (SUM split), we use Rouge-L to measure the overlap between generated summaries and reference summaries.
For the Loong financial split, we follow the official evaluation protocol and use an LLM judge to score model outputs on a scale from 0 to 100.
The judge prompt template is provided in the original paper~\cite{wang-etal-2024-leave} and evaluates responses based on correctness and reasoning quality over financial documents.
We employ DeepSeek-R1-Distill-Qwen-32B~\cite{DeepSeekAI2025DeepSeekR1IR} with thinking enabled and temperature set to 0.0 as judge.

\section{Model Selection}\label{app:model-selection}
To inform our choice of base model for training experiments, we conduct a preliminary evaluation comparing several state-of-the-art long-context language models against their RAG-enhanced counterparts on out-of-domain benchmarks.

\paragraph{Models.}
We evaluate the following models: Qwen2.5-7B-Instruct (128k context), Qwen2.5-7B-Instruct-1M (1M context), glm-4-9b-chat (128k context), glm-4-9b-chat-1m (1M context), and internlm2\_5-7b-chat-1m (1M context)~\cite{qwen2025qwen25technicalreport, glm2024chatglm, cai2024internlm2}.
For RAG baselines, we employ Qwen3-Embedding-4B as the retriever~\cite{zhang2025qwen3embeddingadvancingtext}.
When inputs exceeds the model's supported context window, we apply truncation at the end of the input context.

\paragraph{Benchmarks.}
We assess performance on three long-context benchmarks: $\infty$Bench~\cite{zhang-etal-2024-bench} (MC, QA, and SUM splits), LB-v2~\cite{bai-etal-2025-longbench}, and the English financial split of Loong~\cite{wang-etal-2024-leave}.
These benchmarks provide diverse evaluation scenarios spanning multiple-choice questions, open-ended QA, summarization, and numerical reasoning tasks.

\paragraph{RAG Configuration.}
Following recent findings that advocate for longer chunking strategies with LCLMs~\cite{jiang2024longragenhancingretrievalaugmentedgeneration}, we chunk each document into 2048-token segments.
For each benchmark instance, we retrieve passages to construct contexts of varying lengths $k \in$ \{4k, 8k, 16k, 32k, 64k, 128k, 1M\}, allowing us to analyze performance degradation as context length increases and identify when RAG provides advantages over full-context processing.

\paragraph{Results and Analysis.}

\begin{table*}[t]
\centering
\small
\begin{tabular}{@{}p{0.13\textwidth}p{0.22\textwidth}cp{0.065\textwidth}p{0.063\textwidth}p{0.063\textwidth}p{0.063\textwidth}p{0.075\textwidth}p{0.065\textwidth}@{}}
\toprule
\multirow{2}{*}{\textbf{Setting}} & \multirow{2}{*}{\textbf{Model}} & \textbf{Ctx.} & \textbf{Max.} & \multicolumn{3}{c}{\textbf{$\infty$Bench}} & \textbf{LB-V2} & \textbf{Loong} \\
\cmidrule(lr){5-7} \cmidrule(lr){8-8} \cmidrule(lr){9-9}
& & \textbf{Len.} & \textbf{Input Len.} & \textbf{MC} & \textbf{QA} & \textbf{Sum} & \textbf{ALL} & \textbf{Fin.} \\
\midrule
\multicolumn{9}{@{}l}{\textit{Full Context}} \\
\midrule
& Qwen2.5-7B-Instruct & 128k & 128k & 64.2 & 27.1 & 31.9 & 34.9 & 44.8 \\
& Qwen2.5-7B-Instruct-1M & 1M & 1M & 72.0 & 30.0 & 31.2 & 32.0 & 49.9 \\
& glm-4-9b-chat & 128k & 128k & 70.3 & 36.2 & 27.9 & 29.1 & 49.1 \\
& glm-4-9b-chat-1m & 1M & 1M & 77.3 & 37.3 & 27.7 & 30.4 & 49.6 \\
 & internlm2\_5-7b-chat-1m & 1M & 1M & 71.6 & 36.7 & 20.0 & 26.0 & 32.9 \\
\midrule
\multicolumn{9}{@{}l}{\textit{RAG}} \\
\midrule
& Qwen2.5-7B-Instruct & 128k & 128k & 72.1 & 28.2 & 31.1 & 32.7 & 49.2 \\
& & & 64k & 72.5 & 31.9 & 32.2 & 32.8 & 49.9 \\
& & & 32k & 74.7 & 29.6 & 30.5 & 34.8 & 52.0 \\
& & & 16k & 74.7 & 29.3 & 27.1 & 35.8 & 48.6 \\
& & & 8k & 69.0 & 25.1 & 22.4 & 32.2 & 46.8 \\
& & & 4k & 57.6 & 19.1 & 19.3 & 28.4 & 35.9 \\
\cmidrule(lr){2-9}
& Qwen2.5-7B-Instruct-1M & 1M & 1M & 69.1 & 29.1 & 29.1 & 30.4 & 50.7 \\
& & & 128k & 74.2 & 33.6 & 29.1 & 35.2 & 54.7 \\
& & & 64k & 77.7 & 30.8 & 26.8 & 35.0 & 52.8 \\
& & & 32k & 78.6 & 29.3 & 25.1 & 38.0 & 47.8 \\
& & & 16k & 74.7 & 27.6 & 25.2 & 35.2 & 54.9 \\
& & & 8k & 71.2 & 24.2 & 23.8 & 32.4 & 47.9 \\
& & & 4k & 59.8 & 18.8 & 19.4 & 31.6 & 39.6 \\
\cmidrule(lr){2-9}
& glm-4-9b-chat & 128k & 128k & 75.5 & 33.6 & 26.2 & 31.1 & 50.7 \\
& & & 64k & 74.7 & 32.8 & 23.9 & 31.4 & 48.8 \\
& & & 32k & 76.4 & 29.6 & 22.8 & 30.2 & 45.3 \\
& & & 16k & 71.6 & 27.3 & 20.6 & 30.4 & 43.4 \\
& & & 8k & 68.5 & 21.9 & 19.5 & 30.4 & 43.3 \\
& & & 4k & 54.1 & 15.4 & 15.9 & 29.0 & 31.9 \\
\cmidrule(lr){2-9}
& glm-4-9b-chat-1m & 1M & 1M & 74.7 & 33.3 & 27.2 & 31.8 & 47.6 \\
& & & 128k & 75.5 & 35.3 & 25.9 & 34.0 & 47.3 \\
& & & 64k & 77.3 & 31.3 & 25.1 & 32.6 & 46.1 \\
& & & 32k & 74.7 & 28.2 & 23.1 & 32.8 & 44.7 \\
& & & 16k & 76.4 & 25.1 & 21.6 & 31.8 & 44.1 \\
& & & 8k & 69.4 & 21.9 & 20.1 & 31.6 & 43.1 \\
& & & 4k & 55.0 & 17.1 & 17.3 & 27.2 & 31.7 \\
\cmidrule(lr){2-9}
& internlm2\_5-7b-chat-1m & 1M & 1M & 65.5 & 36.7 & 18.2 & 25.6 & 34.9 \\
& & & 128k & 69.4 & 35.3 & 17.8 & 26.4 & 37.1 \\
& & & 64k & 72.0 & 35.0 & 16.7 & 25.6 & 38.9 \\
& & & 32k & 70.7 & 31.3 & 15.9 & 29.6 & 38.9 \\
& & & 16k & 72.5 & 27.6 & 15.7 & 28.0 & 40.7 \\
& & & 8k & 67.2 & 25.1 & 15.1 & 28.8 & 36.6 \\
& & & 4k & 57.6 & 17.7 & 12.8 & 29.2 & 30.9 \\
\bottomrule
\end{tabular}
\caption{Comparative performance of long-context models in Full Context and RAG settings. MC: Multiple Choice (Accuracy $\uparrow$), QA: Question Answering (Sub-Exact Match $\uparrow$), Sum: Summarization (Rouge-L $\uparrow$), ALL: Overall LB-v2 (Accuracy $\uparrow$), Fin.: Financial split from Loong (LLM Judge Score $\uparrow$). Ctx. Len.: Maximum context window. Max. Input Len.: Maximum number of tokens provided to the model for inference. Bold indicates best performance within each model group.}
\label{tab:model-selection}
\end{table*}

Table~\ref{tab:model-selection} presents the comparative results.
We observe that RAG consistently outperforms full-context processing for most models, with optimal performance typically achieved at moderate retrieval lengths (32k-128k tokens).
However, the results reveal important nuances: while Qwen2.5-7B-Instruct-1M shows substantial improvements with RAG (e.g.,~78.6\% on $\infty$Bench MC with 32k RAG vs. 72.0\% full context), glm-4-9b-chat-1m demonstrates the opposite pattern, achieving its best performance with full-context processing on several tasks (37.3\% on $\infty$Bench QA vs. 35.3\% best RAG performance).
The performance gap is particularly pronounced on the challenging Loong financial split, where RAG configurations frequently outperform full-context processing by 2-5 points for Qwen models.
These mixed results suggest that while some models struggle to effectively attend to relevant information within extended contexts, others have developed stronger selective attention capabilities during pre-training.

Based on these findings, we select Qwen2.5-7B-Instruct-1M as our base model for subsequent training experiments.
This choice is motivated by three key factors: (1) it exhibits the clearest and most consistent performance gap between full-context and RAG settings (6.6 points on $\infty$Bench MC, 3.6 points on $\infty$Bench QA, and 4.8 points on Loong), indicating substantial room for improvement through targeted training on selective attention; (2) unlike glm-4-9b-chat-1m which already demonstrates strong full-context capabilities, Qwen2.5-7B-Instruct-1M represents models that genuinely benefit from retrieval mechanisms, making it an ideal testbed for our hypothesis; and (3) its 1M context window allows us to evaluate whether models trained on moderate-length contexts (16k-32k) can generalize to much longer sequences at inference time.

\section{Detailed Results}\label{app:detailed-results}

\begin{table*}[t]
\centering
\footnotesize
\setlength{\tabcolsep}{3.5pt}
\begin{tabular}{@{}lp{0.24\textwidth}ccccccc@{}}
\toprule
\textbf{Subset} & \textbf{Model} & \textbf{4k} & \textbf{8k} & \textbf{16k} & \textbf{32k} & \textbf{64k} & \textbf{128k} & \textbf{Avg.} \\
\midrule
\multicolumn{9}{@{}l}{\textit{HotpotQA (SubEM $\uparrow$, 300 instances)}} \\
\midrule
& Qwen2.5-7B-Instruct-1M & 70.3 & 68.7 & 64.3 & 57.3 & 53.0 & 52.3 & 61.0 \\
& \quad w/ gold docs & 76.3 & 76.0 & 73.3 & 73.0 & 71.3 & 63.0 & 72.2 \\
& \quad w/o context & 27.0 & 27.0 & 27.0 & 27.0 & 27.0 & 27.0 & 27.0 \\
& + $SFT_{\text{AO}}$ & 62.3 & 54.7 & 55.3 & 52.0 & 45.0 & 39.7 & 51.4 \\
& + $R_{\text{AO}}(y)$ & 85.3 & \textbf{84.3} & \textbf{85.0} & \textbf{80.0} & \textbf{77.7} & \textbf{74.3} & \textbf{81.1} \\
& + $R_{\text{ID}}(y)$ & 83.7 & 81.0 & 77.0 & 75.0 & 70.3 & 64.0 & 75.2 \\
& + $R_{\text{ID+C}}(y)$ & \textbf{86.0} & 74.7 & 69.3 & 60.7 & 60.7 & 56.0 & 67.9 \\
& + $R_{\text{ID+Q}}(y)$) & 84.7 & 80.0 & 76.3 & 74.0 & 69.7 & 60.0 & 74.1 \\
& + $R_{\text{R+Judge}}(y)$ & 84.3 & 78.0 & 75.3 & 66.7 & 62.3 & 57.0 & 70.6 \\
\midrule
\multicolumn{9}{@{}l}{\textit{NaturalQuestions (SubEM $\uparrow$, 600 instances)}} \\
\midrule
& Qwen2.5-7B-Instruct-1M & 63.7 & 63.7 & 61.5 & 60.7 & 60.0 & 57.0 & 61.1 \\
& \quad w/ gold docs & 73.0 & 74.7 & 73.8 & 71.5 & 66.8 & 64.3 & 70.7 \\
& \quad w/o context & 32.5 & 32.5 & 32.5 & 32.5 & 32.5 & 32.5 & 32.5 \\
& + $SFT_{\text{AO}}$ & 48.3 & 41.5 & 41.0 & 38.7 & 34.2 & 37.5 & 40.3 \\
& + $R_{\text{AO}}(y)$ & 70.2 & 69.2 & 67.7 & 67.5 & 66.8 & \textbf{64.8} & 67.7 \\
& + $R_{\text{ID}}(y)$ & 70.3 & 66.8 & 66.3 & 65.8 & 62.2 & 61.5 & 65.5 \\
& + $R_{\text{ID+C}}(y)$ & 66.2 & 62.7 & 55.2 & 55.5 & 52.7 & 49.2 & 56.9 \\
& + $R_{\text{ID+Q}}(y)$ & 71.2 & 63.5 & 59.8 & 58.3 & 53.7 & 48.2 & 59.1 \\
& + $R_{\text{R+Judge}}(y)$ & \textbf{80.8} & \textbf{77.3} & \textbf{74.7} & \textbf{72.0} & \textbf{67.2} & 64.7 & \textbf{72.8} \\
\midrule
\multicolumn{9}{@{}l}{\textit{TriviaQA (SubEM $\uparrow$, 600 instances)}} \\
\midrule
& Qwen2.5-7B-Instruct-1M & 84.8 & 86.8 & 87.2 & 87.7 & 87.2 & 85.8 & 86.6 \\
& \quad w/ gold docs & 87.3 & 88.7 & 88.2 & 87.2 & 87.3 & 76.2 & 85.8 \\
& \quad w/o context & 56.2 & 56.2 & 56.2 & 56.2 & 56.2 & 56.2 & 56.2 \\
& + $SFT_{\text{AO}}$ & 83.3 & 77.2 & 76.2 & 74.0 & 73.5 & 73.2 & 76.1 \\
& + $R_{\text{AO}}(y)$ & 89.2 & 90.0 & \textbf{92.7} & \textbf{93.2} & \textbf{93.7} & \textbf{93.5} & 92.1 \\
& + $R_{\text{ID}}(y)$ & 89.8 & 91.8 & 91.2 & 92.3 & 90.3 & 90.3 & 91.0 \\
& + $R_{\text{ID+C}}(y)$ & 90.2 & 86.3 & 85.5 & 84.3 & 85.2 & 83.5 & 85.8 \\
& + $R_{\text{ID+Q}}(y)$ & 87.8 & 86.5 & 83.0 & 83.7 & 86.7 & 85.0 & 85.5 \\
& + $R_{\text{R+Judge}}(y)$ & \textbf{93.5} & \textbf{93.2} & 92.0 & 92.7 & 92.3 & 89.8 & \textbf{92.3} \\
\bottomrule
\end{tabular}
\caption{Performance on HELMET retrieval subsets across varying context lengths. Base: Qwen2.5-7B-Instruct-1M. "w/ gold docs" gives the model explicit information about which documents are relevant; "w/o context" provides no context (evaluates parametric knowledge). Bold indicates best performance per column.}
\label{tab:in-domain-results-detailed}
\end{table*}

\begin{table*}[t]
\centering
\small
\setlength{\tabcolsep}{4pt}
\begin{tabular}{@{}llccccccc@{}}
\toprule
\multirow{2}{*}{\textbf{Setting}} & \multirow{2}{*}{\textbf{Model}} & \multirow{2}{*}{\textbf{Max Len}} & \multicolumn{3}{c}{\textbf{$\infty$Bench}} & \textbf{LB-v2} & \textbf{Loong} \\
\cmidrule(lr){4-6} \cmidrule(lr){7-7} \cmidrule(lr){8-8}
& & & \textbf{MC} & \textbf{QA} & \textbf{Sum} & \textbf{ALL} & \textbf{Fin.} \\
& & & \scriptsize{(Acc. $\uparrow$)} & \scriptsize{(SubEM $\uparrow$)} & \scriptsize{(R-L $\uparrow$)} & \scriptsize{(Acc. $\uparrow$)} & \scriptsize{(Judge $\uparrow$)} & \\
\midrule
\multicolumn{8}{@{}l}{\textit{Full Context (unaltered)}} \\
\midrule
& Qwen2.5-7B-Instruct-1M & 1M & 72.0 & 30.0 & 31.2 & 32.0 & 49.9 \\
& \quad + $R_{\text{AO}}(y)$ & 1M & 70.3 & \textbf{38.2} & 31.0 & 31.2 & 55.3 \\
& \quad + $R_{\text{ID}}(y)$ & 1M & 71.6 & 34.7 & 29.7 & 31.0 & 51.2 \\
& \quad + $R_{\text{ID+C}}(y)$ & 1M & 70.7 & 31.3 & 30.2 & 31.4 & 47.5 \\
& \quad + $R_{\text{ID+Q}}(y)$ & 1M & 70.7 & 30.8 & 30.5 & 32.0 & 51.5 \\
& \quad + $R_{\text{R+Judge}}(y)$ & 1M & 72.5 & 30.7 & \textbf{31.7} & 32.6 & \textbf{58.8} \\
\midrule
\multicolumn{8}{@{}l}{\textit{KV-Cache with RetrievalAttn}} \\
\midrule
& Qwen2.5-7B-Instruct-1M & 1M & 62.0 & 27.6 & 27.3 & 30.2 & 39.5 \\
& \quad + $R_{\text{AO}}(y)$ & 1M & 65.5 & 33.6 & 28.0 & 29.6 & 41.7 \\
& \quad + $R_{\text{ID}}(y)$ & 1M & 64.2 & 29.3 & 26.0 & 30.0 & 43.0 \\
& \quad + $R_{\text{ID+C}}(y)$ & 1M & 63.3 & 27.1 & 27.7 & 29.6 & 39.8 \\
& \quad + $R_{\text{ID+Q}}(y)$ & 1M & 62.9 & 27.1 & 28.1 & 32.0 & 39.7 \\
& \quad + $R_{\text{R+Judge}}(y)$ & 1M & 63.8 & 28.2 & 28.8 & 31.2 & 41.5 \\
\midrule
\multicolumn{8}{@{}l}{\textit{RAG}} \\
\midrule
& Qwen2.5-7B-Instruct-1M & 1M & 69.1 & 29.0 & 29.1 & 30.4 & 50.7 \\
& & 128k & 74.2 & 33.6 & 29.1 & 35.2 & 54.7 \\
& & 64k & 77.7 & 31.0 & 26.8 & 35.0 & 52.8 \\
& & 32k & \textbf{78.6} & 29.3 & 25.1 & \textbf{38.0} & 47.8 \\
& & 16k & 74.7 & 27.6 & 25.2 & 35.2 & 54.9 \\
& & 8k & 71.2 & 24.2 & 23.8 & 32.4 & 47.9 \\
& & 4k & 59.8 & 18.8 & 19.4 & 31.6 & 39.6 \\
\bottomrule
\end{tabular}
\caption{Performance comparison across different training strategies and inference settings. Numbers in parentheses indicate the number of test instances. Bold values indicate best performance within each major setting group. MC: Multiple Choice (Accuracy $\uparrow$), QA: Question Answering (SubEM $\uparrow$), Sum: Summarization (Rouge-L $\uparrow$), ALL: Overall (Accuracy $\uparrow$), Fin.: Financial (Judge Score $\uparrow$).}
\label{tab:ood-results-detailed}
\end{table*}

\paragraph{In-Domain Results.}\label{app:results-id}

Table~\ref{tab:in-domain-results-detailed} presents performance on HELMET's retrieval subsets across varying context lengths. 
From the results we can immediately see how standard SFT leads to severe performance degradation compared to the base model, while RL-based approaches consistently outperform the baseline.
The $R_{\text{AO}}(y)$ reward function achieves the strongest and most consistent improvements, with average gains of 20.1 points on HotpotQA (81.1 vs. 61.0), 6.6 points on NQ (67.7 vs. 61.1), and 5.5 points on TriviaQA (92.1 vs. 86.6) over the base model. 
Notably, $R_{\text{AO}}(y)$ demonstrates superior robustness to context length scaling, maintaining 74.3\% accuracy on HotpotQA at 128k tokens compared to 52.3\% for the base model—a 22-point advantage at the longest context length.

The $R_{\text{ID}}(y)$ reward function also yields substantial gains, averaging 75.2 on HotpotQA and 91.0 on TriviaQA. 
However, strategies requiring more constrained outputs—specifically $R_{\text{ID+C}}(y)$ and $R_{\text{ID+Q}}(y)$—show marked performance degradation as context length increases. 
For instance, $R_{\text{ID+C}}(y)$ achieves 90.2\% at 4k on TriviaQA but drops to 83.5\% at 128k, performing worse than the base model at longer contexts. 
This pattern suggests that overly restrictive training objectives may impede the model's ability to process extended contexts effectively.

Comparing against the upper bound (w/ gold docs) and lower bound (w/o context), we observe that fine-tuned models not only substantially improve over the base, but they also surpass the performance obtained when the relevant documents are provided, particularly on HotpotQA (81.1 vs. 72.2 gold docs) and TriviaQA (92.1 vs. 85.8 gold docs), indicating that the training successfully enables selective attention without sacrificing the model's ability to leverage its full context when beneficial.

\paragraph{Out-of-Domain Results.}\label{app:results-ood}

Table~\ref{tab:ood-results-detailed} evaluates generalization to out-of-domain benchmarks where inputs consist of long, cohesive documents rather than retrieved passages. Results reveal mixed patterns across tasks. 
The $R_{\text{R+Judge}}(y)$ approach demonstrates the most substantial improvement on the Loong financial split, achieving 58.8 (+8.9 over base), while $R_{\text{AO}}(y)$ yields the largest gain on $\infty$Bench QA (38.2, +8.2 points). 
However, improvements are modest or absent on other benchmarks: multiple-choice accuracy remains largely unchanged (70.3--72.5 vs. 72.0 base), and summarization shows minimal variation (29.7--31.7 vs. 31.2 base).

Notably, RAG baselines at moderate retrieval lengths (32k--128k tokens) frequently outperform all fine-tuned full-context models. 
For instance, RAG@32k achieves 78.6\% on $\infty$Bench MC compared to 72.5\% for the best fine-tuned model ($R_{\text{R+Judge}}(y)$).
This gap is particularly pronounced on LB-v2, where RAG@32k scores 38.0\% versus 32.6\% for the best full-context model. 
These results suggest that while fine-tuning improves selective attention for in-domain retrieval scenarios, it provides limited advantages for tasks requiring holistic document understanding—domains where RAG's filtering mechanism retains its efficacy.

\end{document}